\documentclass[11pt, a4paper, logo, copyright, singlecolumn]{deepmind}
\pdfoutput=1

\pdftrailerid{redacted}

\makeatletter
\renewcommand\bibentry[1]{\nocite{#1}{\frenchspacing\@nameuse{BR@r@#1\@extra@b@citeb}}}
\makeatother

\usepackage{dsfont}
\usepackage{dm-colors}
\usepackage{xcolor}
\usepackage{graphicx}
\usepackage{hyperref}

\usepackage{amsmath,amssymb}
\newcommand{\myvecsym}[1]{\boldsymbol{#1}}
\newcommand{\vtheta}{\myvecsym{\theta}}

\newcommand{\agent}{MIA}

\usepackage[authoryear,round]{natbib} 

\title{Creating Multimodal Interactive Agents with Imitation and Self-Supervised Learning}
\author[1]{Interactive Agents Team}
\affil[1]{DeepMind}

\renewcommand{\today}{}

\usepackage{graphicx}

\begin{abstract}
A common vision from science fiction is that robots will one day inhabit our physical spaces, sense the world as we do, assist our physical labours, and communicate with us through natural language. Here we study how to design artificial agents that can interact naturally with humans using the simplification of a virtual environment. We show that imitation learning of human-human interactions in a simulated world, in conjunction with self-supervised learning, is sufficient to produce a multimodal interactive agent, which we call \agent{}, that successfully interacts with non-adversarial humans 75\% of the time. We further identify architectural and algorithmic techniques that improve performance, such as hierarchical action selection. Altogether, our results demonstrate that imitation of multi-modal, real-time human behaviour may provide a straightforward and surprisingly effective means of imbuing agents with a rich behavioural prior from which agents might then be fine-tuned for specific purposes, thus laying a foundation for training capable agents for interactive robots or digital assistants. A video of \agent{}'s behaviour may be found at \url{https://youtu.be/ZFgRhviF7mY}.
\end{abstract}

\begin{document}
\maketitle

\section{Introduction}

Humans interact with the physical world and one another, and it is through these interactions that much of our cognition was shaped during evolution~\citep{dunbar1993coevolution}. If we hope to build artificial intelligence (AI) capable of human-like thinking, we should therefore consider a holistic setting wherein agents perceive and manipulate their world and understand and produce language, so that they can participate in---and learn from---natural interactions with humans~\citep{mcclelland2019extending,lake2021word,winograd1972understanding}. 

In this work we explore how to create artificial agents that interact with humans and their environment. To do so, we use  imitation learning approaches~\citep{pomerleau1989alvinn,schaal1999imitation} that have driven progress in Go \citep{silver2016mastering}, Starcraft \citep{vinyals2019grandmaster}, and perhaps most notably the recent progress in large language models \citep{brown2020language} and dialogue agents \cite{adiwardana2020towards}. Imitation learning has proved a surprisingly powerful approach for games and language modelling, but it is unclear the extent to which it may furnish powerful behavioural priors in embodied domains. We study this question in a 3D simulator that is easily operated by human participants. Working to create embodied agents with impressive motor and linguistic capabilities is a difficult problem for several practical reasons, notably: (1) model requirements are more sophisticated, and (2) data sources are not widely available, as they are with text. Our results suggest, however, that by augmenting imitation learning with hierarchical architectures and self-supervised learning it is possible to make agents that are capable, impressive, and surprising using moderately-sized datasets collected directly by researchers.

We build upon our previously introduced methodology~\citep{abramson2020imitating} in a few crucial ways. First, we significantly increase the scale and complexity of collected data. Second, we simplify the training protocol. Third, we remove all uses of privileged information extracted from our virtual 3D-world simulator. Ultimately, these changes result in an agent artifact that blends perception, language understanding and production, and motor action to competently engage in extended, and often surprising interactions, and participate in comprehensive, bidirectional language-based communication. The resulting agent, \agent{} (\textbf{M}ultimodal \textbf{I}nteractive \textbf{A}gent), is vastly more capable than those presented in the previous work~\citep{abramson2020imitating}. \agent{} exhibits a diversity of behaviours that were never instructed by researchers, including tidying a room and finding and grouping multiple specified objects. It asks clarifying questions, produces minute long action sequences following coherent goals, and can rapidly learn new objects, nouns, and verbs in mere hours of real-time experience. 

\noindent
Our contributions are as follows:
\vspace{-3mm}
\begin{itemize}
    \item We introduce the multi-room Playhouse environment to study natural interactions between humans and agents.
    \item We produce an agent, \agent{}, which is capable of engaging in natural interactions with human participants in a 3D simulated world.
    \item We demonstrate the importance of architectural design and self-supervised losses to performance in situated language agents domains.
    \item We highlight the importance of human and programmatic evaluation beyond training and validation losses.
    \item We demonstrate the effects of scale of both data and model size on performance.
    \item We demonstrate the importance of large-scale data priors for fast learning of new objects and skills as well as their referring nouns and verbs in a practical amount of time (hours).
    \item We provide a careful comparison to human performance, as evaluated by humans, on tasks involving natural interactions.
\end{itemize}

\section{Language Games in the Playhouse}
\begin{figure}[ht]
    \centering
    \includegraphics[width=\textwidth]{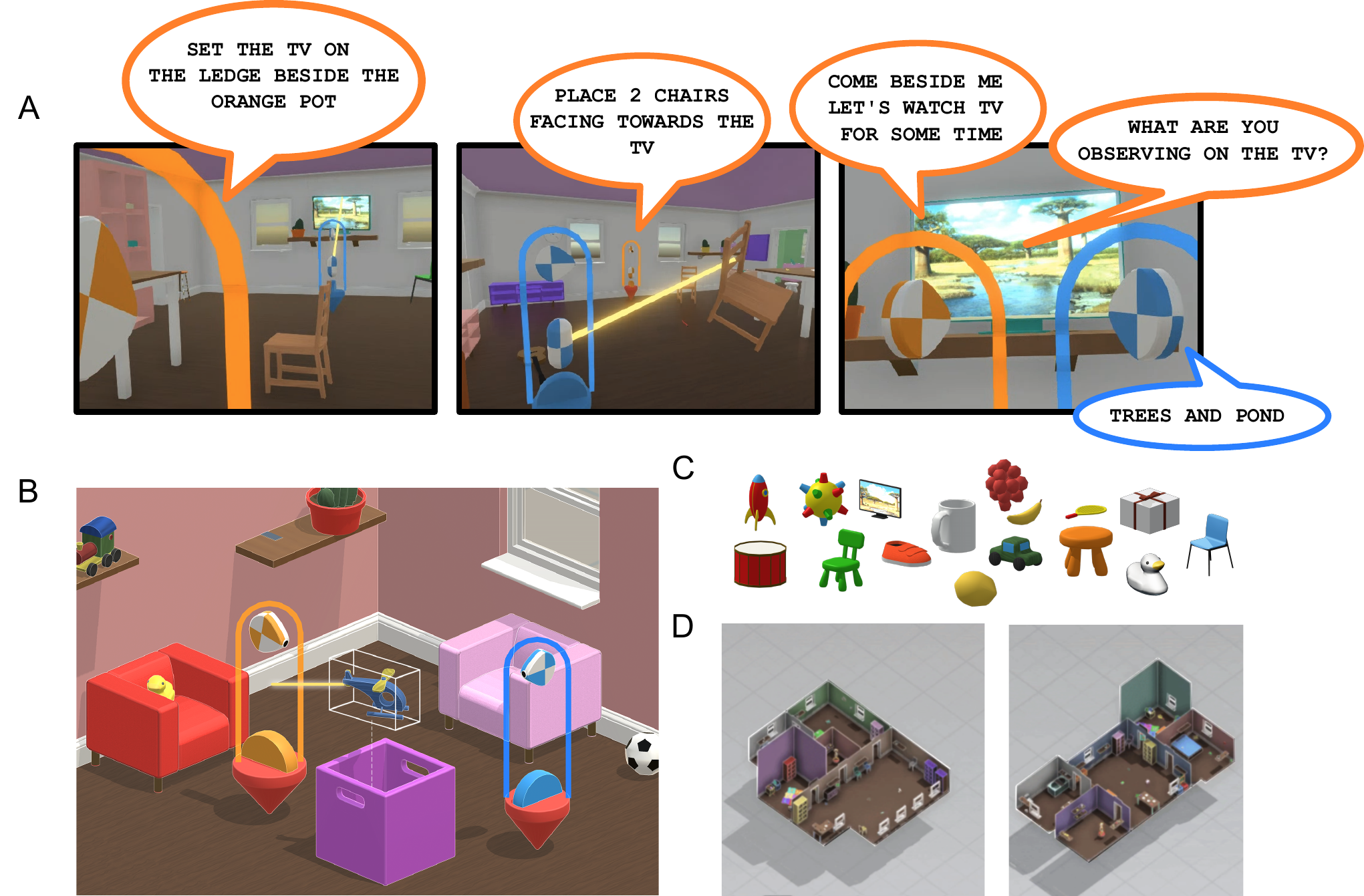}
    \caption{\textbf{Interactions in a simulated “Playhouse” environment.} Humans and agents interact via simulated avatars in the 3-D “Playhouse” environment. The environment contains a randomised set of rooms with domestic objects and children's toys, as well as containers, shelves, furniture, windows, and doors. The diversity of the environment enables interactions involving reasoning about space and object relations, ambiguity of references, containment, construction, support, occlusion, and partial observability. Agents interact with the world by moving around, manipulating objects, and speaking to each other. \textbf{A} Example of two humans interacting in the Playhouse. This episode was selected from a small set that was drawn at random from our dataset.  \textbf{B} Depicts a simple interaction wherein the orange solver agent is placing a helicopter into a container while the blue setter agent watches on. \textbf{C} A sampling of the types of objects available in the room. \textbf{D} Two random instantiations of the Playhouse, each with unique configurations of rooms, furniture, and objects.   
    }
    \label{fig:1}
\end{figure}

We explored human and agent interaction in a 3D virtual environment called the Playhouse, based on previous environments built in the Unity game engine~\citep{ward2020using,abramson2020imitating}. The Playhouse comprises a randomized set of adjoining rooms (such as a living room, pantry, and bathroom) and interactable items (such as toys and household objects). Human and agent embodiment manifests as control of a virtual robot that moves around in space, grasps and manipulates objects, and emits natural language. 
This setting permits a wide range of behaviours, ranging from simple instruction-following (e.g., ``\emph{Please pick up the book from the floor and place it on the blue bookshelf}.'') to creative acting (e.g., \emph{``Bring food to the table so that we can eat}''). 

It is not feasible to programmatically define reward functions for the Playhouse because the space of all possible interactions is vast and unpredictable, and because satisfaction criteria can often be ambiguous or subjective. We might instead choose to task humans with assigning reward based on their subjective assessment of an agent's behaviour, either by absolute score or ranking~\citep{christiano2017deep}, following an interaction. However, unless these interactions could occur at an enormous scale and fast enough pace, this too is not a feasible approach to training initially randomly behaving agents using purely reinforcement. 

Therefore, we follow an approach advocated for by~\citep{abramson2020imitating} wherein we train agents using imitation to instill them with an intelligent \emph{behavioural prior}~\citep{galashov2019information}. Agents trained in this manner will display some structured and language-driven behaviours. Subsequent reinforcement learning will be more efficient with these agents since humans will be more inclined to continue interactions with AI participants that display rudimentary intelligence. The focus of this report centers on creating agents with intelligent behavioural priors, and we leave further reward-based learning for future work.

\subsection{Collecting Data in the Playhouse}
Imitation learning works best when demonstrations are produced by experts and when there are sufficiently many to cover the desired range of behaviours (including, e.g., corrective behaviours~\citep{ross2011reduction}). For domains such as language, data can be easily obtained by leveraging existing text on the internet~\citep{brown2020language}. However, for domains such as robotics (or simulated robotics), data needs to be assembled from scratch. The protocol for organizing the collection of expert demonstration data in the Playhouse is a central contribution of this work.

One approach could be to collect data from humans freely interacting in the Playhouse. However, such data might prove problematic for a host of reasons. First, data may be difficult to model as it might not comprise graded behavioural competencies. Humans come to the environment with an ability to act in simulated worlds, an extensive knowledge base, and nuanced intentions, and hence exhibit already-complex behaviour that takes for granted simpler, but necessary, capacities such as the ability to identify basic objects. Second, as interactions are guided by the whims of human participants, collected behaviours might naturally collapse to a few characteristic specific modes, and hence might not span the behaviours we ultimately care about. Third, if left unchecked the data might be permeated with undesirable biases. A different approach could be to collect data using templated scripts that human experts must follow. However, this places a burden on researchers to infer the set of scripts whose translations span the desired range of behaviours, which is an impossible feat if one wishes to capture the contextual, ambiguous, and nuanced forces that drive natural human behaviour. 

Our approach occupies a middle ground between these two possibilities. We centered Playhouse interactions on \textit{language games}, which are a collection of both prompted interaction-types and free-form instructions~\citep{lynch2020grounding} from which human players can base their behaviour~\citep{abramson2020imitating}. All language games had the same basic structure: a \textit{setter} would receive a prompt from which they could design an interaction, which they propose to a \textit{solver} agent. For example, the setter might receive a prompt to ``Ask the other player a question about the existence of an object'', and after some exploration to discover possibilities, the setter could translate this into the task: ``Please tell me whether there is a blue duck in a room that does not also have any furniture''. The combination of (1) a small set of prompts, (2) the random, combinatoric structure of the Playhouse, and (3) the natural linguistic and intentional variation that humans exhibit ensured that interactions were unique, and together spanned the set of competencies we hoped agents to learn. To ensure we sufficiently captured the richness of natural human behaviour, we also included free-form instructions, which gave setters more leeway in guiding interactions, within certain constraints (namely, interactions should resemble, in the participants' best judgement, those from the templates, and should be closely monitored for any arising ethical concerns). A notable example of creative free-form instructions is: ``Now take any object that you like and hit the tennis ball off the stool so that it rolls near the clock, or somewhere near it.'' 

\begin{table}
\begin{center}
\begin{tabular}{ ccccc }
 & Totals & & & \\
 \hline
 Number of train episodes & 275676 & & & \\
 Number of test episodes & 34399 & & & \\
 Episode length (minutes) & 5 & & & \\
 Total play time (years) & 2.94 & & & \\
 Number of unique setter instructions & 778808 & & & \\
 \hline
 \hline
 & Mean & Std & 25th & 75th \\
 \hline
 Interaction time per instruction (seconds) & 38.86 & 29.93 & 18.00 & 49.00 \\ 
 Setter utterance length & 7.88 & 3.37 & 6.00 & 10.00 \\ 
 Solver utterance length & 2.08 & 2.07 & 1.00 & 2.00 \\
 Number of setter instructions per episode & 3.98 & 1.80 & 3.00 & 5.00 \\
 Idling proportion & 0.69 & 0.26 & 0.50 & 0.91 \\
 \hline
\end{tabular}
\end{center}
\caption{\centering \label{tab:data-stats} \textbf{Playhouse dataset statistics.}}
\end{table}

\section{Multimodal Interactive Agent Training \& Design}

An abundance of expert human behavioural data and a lack of programmatic reward functions motivate a straightforward training regime centered on basic representation learning techniques, which has seen much success in recent large-scale models such as GPT-3~\citep{brown2020language}. Below we describe the essential ingredients: supervised learning of actions~\citep{pomerleau1989alvinn,osa2018algorithmic}, and an additional cross-modal self-supervised learning objective that significantly improves performance beyond supervised learning alone. Since trained agents are difficult to evaluate given the free-form nature of the Playhouse environment, we also describe a protocol for assessing learned behaviour relative to human baselines.

\subsection{Training}
\subsubsection{Behavioural Cloning}
Training comprises a straightforward imitation learning technique, behavioural cloning (BC), which frames behavioural copying as a sequence learning problem. Our data comprise temporal sequences of human behaviour, called trajectories, which include observation sequences $\mathbf{o}_{\leq T} \equiv (\mathbf{o}_0, \mathbf{o}_1, \mathbf{o}_2, \dots, \mathbf{o}_T)$ and action sequences $\mathbf{a}_{\leq T} \equiv (\mathbf{a}_0, \mathbf{a}_1, \mathbf{a}_2, \dots, \mathbf{a}_T)$. Given a dataset of solver-only trajectories we implement the following loss function, which we derive from the forward Kullbach-Leibler divergence between the expert and agent policies in~\cite{abramson2020imitating}:
\begin{align*}
\mathcal{L}^\textsc{bc}(\vtheta) & = -\frac{1}{B} \sum_{n=1}^{B} \sum_{t=0}^{K} \ln \pi_{\vtheta}(\mathbf{a}_{n,t} \mid \mathbf{o}_{n,\leq t}),
\end{align*} 
where $B$ is the minibatch size, $K$ is the backpropagation-through-time window size.

\subsubsection{Modality Matching using Contrastive Self-Supervised Learning}
Techniques that improve generalization will benefit agents in the Playhouse because evaluation environments are bound to be unique from the agent's training data. This is due to the combinatorics and programmatic randomness of the Playhouse (variations in house layouts, object existence, and object placement can produce an inexhaustible source of newness), and because of the dependence of the agent's observations on its actions (deviations in observations or actions from that which are exactly specified by the data will drive the agent to never-before-seen states).

We found one contrastive self-supervised representation learning technique, which involves cross-modality matching \citep{alayrac2020self}, particularly useful in this regard. It is implemented as an auxiliary training loss wherein the agent must predict whether vision and language embeddings match (i.e., they are produced from a trajectory from the dataset as normal), or they do not match (i.e., visual embeddings are produced from the input image of one trajectory in the dataset, and language embeddings are produced from the language input from a different trajectory). The agent processes these visual and language embeddings using its perceptual encoder (see section \ref{sec:perception}), the output of which is fed to an MLP discriminator that produces a binary predictor of whether the embeddings match or not. Practically, we implement this at the level of the minibatch: the batch of data used for the behavioural cloning loss comprise the ``matches'', and a shuffling of vision and language embeddings from the minibatch comprise the mis-matches:
\begin{align}
\mathcal{L}^\textsc{CR}(\vtheta) & = 
-\frac{1}{B} \sum_{n=1}^{B} \sum_{t=0}^T  \bigg [ \ln D_{\vtheta}(\mathbf{o}_{n,t}^\textsc{V}, \mathbf{o}_{n,t}^\textsc{L}) + \ln \big( 1 - D_{\vtheta}(\mathbf{o}_{n,t}^\textsc{V}, \mathbf{o}_{\textsc{Shift}(n),t}^\textsc{L}) \big ) \bigg ],
\end{align}
where $B$ is the batch size and $\textsc{Shift}(n)$ is the $n$-th index after a modular shift of the integers: $1 \to 2, 2 \to 3 \dots, B \to 1$, and superscripts denote the modality ($V$ for vision, $L$ for language). Although language emissions are sparse in each trajectory, we implement language input observations as ``sticky''; e.g., upon the setter emitting a language utterance, the solver will repeatedly receive this same utterance as input until the setter emits a new utterance. This implies that there will always be vision and language pairs that can be matched (or not) at each timestep, except for the first timesteps in an episode that occur prior to the setter speaking. For simplicity, we use $D_{\vtheta}()$ to denote the processing of observations up through the agent's perceptual encoder, and into a separate discriminator MLP, which is only engaged during learning. While the addition of this loss requires an extra forward pass through the perceptual encoder, we observe that the computational overhead results in less than a $1\%$ deduction in training speed.

\begin{figure}[h]
    \centering
    \includegraphics{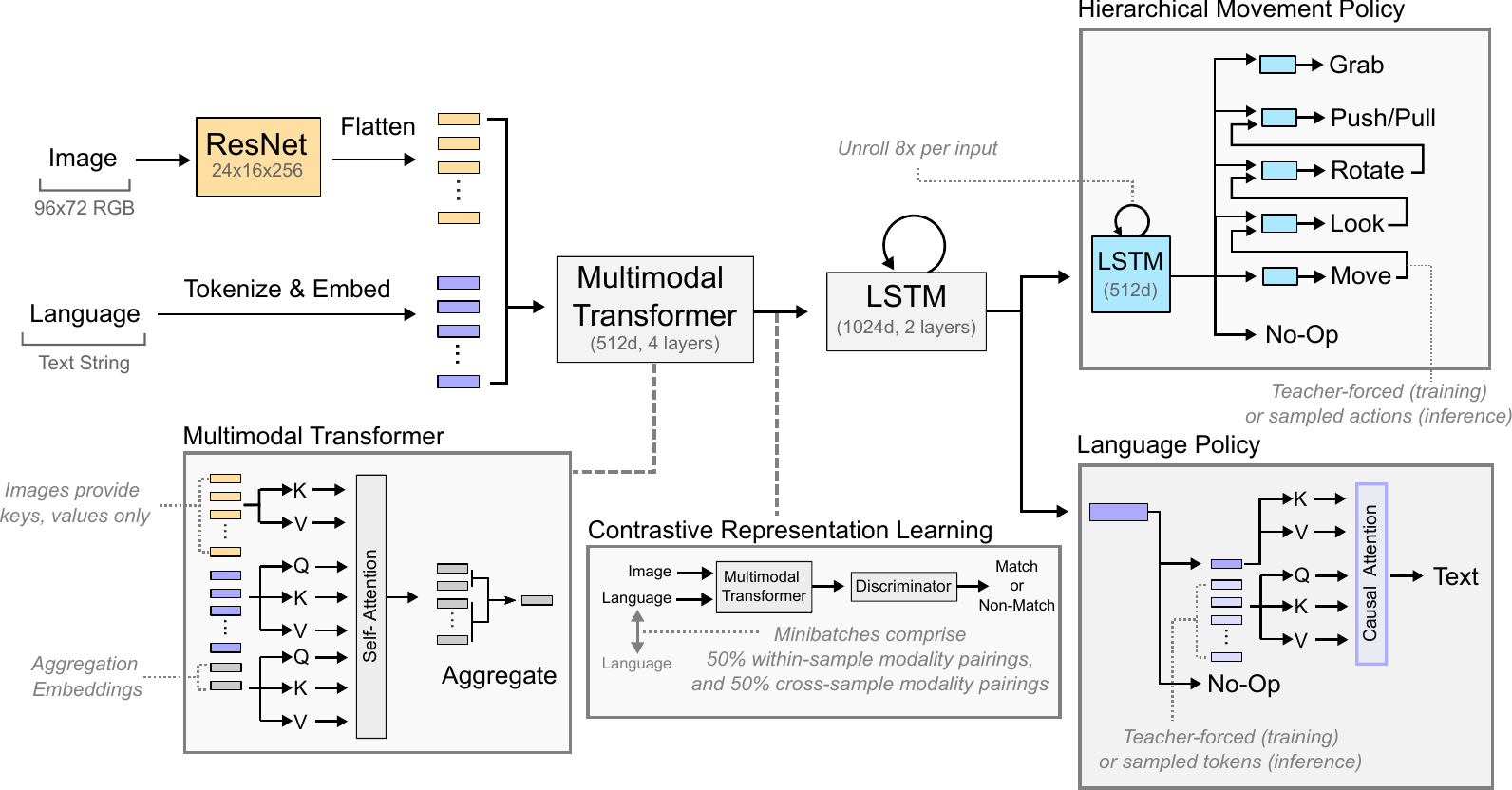}
    \caption{\textbf{Multimodal Interactive Agent (\agent{}) Design.} A series of ResNet blocks downsample the incoming image, while language tokens index an learnable embedding table. Together these embeddings comprise the input to a multi-modal Transformer, whose output is aggregated and provided as input to an LSTM memory. The output of the LSTM conditions both the hierarchical movement policy and language policy, implemented as an LSTM (which unrolls $8\times$ per input, producing $8$ sets of consecutive movement actions) and a Transformer, respectively. \agent{} is trained using behavioural cloning on each of its actions (no-op, move, look, rotate, push and pull, grab, and text) in addition to an auxiliary contrastive loss. Please see the main text for full details.}
    \label{fig:2}
\end{figure}

\subsection{Perception}
\label{sec:perception}
\agent{} receives input from visual ($96\times72$ RGB pixels) and symbolic (text-based language tokens) modalities. For vision we employ conventional ResNet blocks ~\citep{he2016deep} (with strides $(1, 1, 2, 2)$ per block and a consistent kernel shape of $(3,3)$, altogether comprising $20$ total convolution layers), which produce a $24 \times 18 \times 512$ output representation from the input image. For language processing, we produce embeddings from a tokenized version of language input by indexing a learnable embedding table with a vocabulary of $4000$ subwords. The $432$ ($24\times18$) vectors from the visual spatial array, the language embeddings, and two dedicated output embeddings (akin to the ``CLS''embeddings in the BERT model~\citep{devlin2018bert}) are gathered to form the input to a multi-modal transformer (MMT)~\citep{vaswani2017attention}, consisting of $4$ layers, $8$ heads, and $512$ embedding size. The output of the MMT is the concatenation of the two output embeddings and a feature-wise mean pooling of all other output embeddings (that is, all the embeddings that produce queries, which include the two dedicated output embeddings and the language embeddings. The visual embeddings are only cross-attended to, and hence only produce keys and values). This is then provided as input to a two-layer, $1024$-unit LSTM ``memory'', whose output is fed to each policy head.

\subsection{Hierarchical Control}

\subsubsection{High-Level Control}
In the non-hierarchical setting, \agent{} receives observations and produces a set of movement actions $15$ times per second. In the hierarchical setting, new observations arrive only every $8$ steps (i.e., $3.75$ times per second), and thus, \agent{} is tasked with producing $8$ consecutive sets of movement actions in a row before receiving a new observation. Hierarchical control is implemented with an LSTM that receives input from the ``memory'' LSTM, and unrolls for $8$ steps, providing inputs to each of the movement policies (see section \ref{sec:low-level-control}) on each of these $8$ ``internal'' steps. Language actions are emitted from the high-level controller. They're sampled one token at-a-time, up to a maximum of $25$ per high-level step, using the same vocabulary as for language-input tokenization. For the language policy we use a transformer with teacher forcing ($4$ layers, $8$ heads, and $256$ embedding size). The language policy, like the movement policies (see section \ref{sec:low-level-control}) also models "no-ops", which are binary decisions to predict a non-action (during training) or produce a non-action (during acting).

\subsubsection{Low-Level Control}
\label{sec:low-level-control}
\agent{}'s movement action space largely resembles that used by humans: \emph{Look} actions are modelled as a mouse movement to a bin in visual space ($(-1, 1)$ for $x-$ and $y-$ coordinates, using $51$ bins for each), implemented as two MLPs. \emph{Move} actions are similarly modelled as a decision in a binned, two-dimensional translational-force space ($(-1, 1)$ for $up-down$ and $left-right$ translations, using $101$ bins for each), implemented as a MLPs. Object rotations (i.e., angular movements applied to objects when they are being grasped) are modelled along the three cardinal axes, implemented as a recursive discrete decision procedure whereby the agent first chooses from among $3$ coarse bins between $(-1, 1)$, and then chooses from $3$ finer bins within the previously chosen coarse bin, and so on, until a choice is made at a granularity of $0.1$. This is implemented using an LSTM. Push and pull actions are similarly modeled. \emph{Grab} actions are modelled as a binary action (e.g., mouse click), implemented as an MLP.  We also found a form of autoregressivity to be useful: within each policy, look actions in the $x-$ dimension condition look actions in the $y-$ dimension. Rotations in the $x-$ and $y-$ dimensions condition rotation in the $z-$ dimension. Between policies, move actions condition look actions, and move and look actions condition rotation actions, and move, look, and rotation actions condition push and pull actions. There is no autoregressivity across time.

\subsection{Evaluation}
We monitored \agent{}'s performance in three ways: (1) via the total loss and its sub-components (see Figure~\ref{fig:2}A), (2) via scripted probe tasks that serve as a heuristic evaluation of rolled-out behaviours during training (Figure~\ref{fig:2}C), and (3) via live human-agent interactions (Figure~\ref{fig:2}B). In some domains, like language, log probabilities correlate well with model ``performance''. However, in our setting this proved to not be true; e.g., drastically overfit models can perform better, and architecture changes that improve log probabilities can result in worse performance. Therefore, scripted probe tasks and live human-agent interactions are central to agent iteration, despite being more computationally demanding, slow, and subjective.

\subsubsection{Scripted Probe Tasks}
We constructed a suite of probe tasks to evaluate performance on several dimensions, including counting, identifying colors, lifting items, and more refined object positioning. For each task, agents were presented with a templated instruction (e.g., ``place a duck on top of the table''), which was then evaluated using a scripted reward model. Crucially, these reward models were used for evaluation only; agents never received rewards from the environment for the purposes of optimization. 

\subsubsection{Human Evaluation}
Since we aim to build agents that can carry our natural interactions with humans, the ultimate metric we care about is human evaluation of agent interactions. So, upon training agents we asked human participants to interact with agents in a setting that closely mimicked data collection. Humans played the role of the setter, while agents performed the role of the solver. Importantly, humans were not adversarial: they did not search for agent failures, nor did they try to coax the agent into undesirable modes of behaviour. Rather, they carried out interactions similar to those as when they interacted with humans during data collection. After an interaction, human setters provided feedback as to whether the interaction was successful, which we then report as the agent success rate.
\section{Performance, Ablations, \& Modifications}

\begin{figure}[ht]
    \centering
    \includegraphics[width=\textwidth]{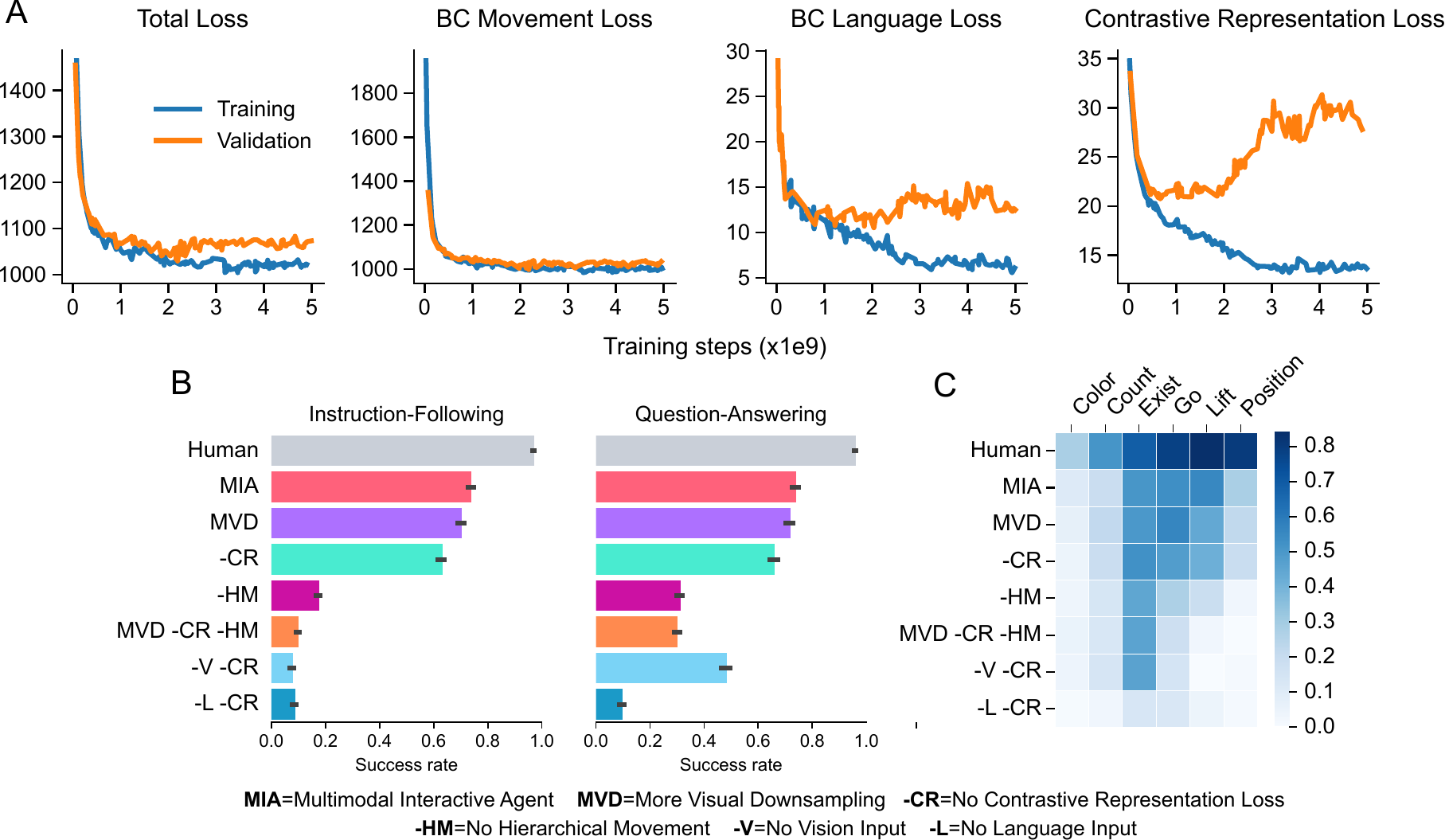}
    \caption{\textbf{Performance, ablations and modifications.} \textbf{A} Loss curves. The top row shows the loss of \agent{} on training and validation datasets during training. After training on 5G timesteps the behavioural cloning loss on movement actions is fairly flat while the behavioural cloning loss on language actions and the contrastive representation learning loss have started to overfit.
 \textbf{B} Human evaluation. The bottom left plot compares the success rate of humans and \agent{}, with various ablations applied, when playing the role of a solver in a Playhouse language game, as judged by human setters. The episodes are divided according to the prompt given to the human setter, as either ``Instruction following'', e.g. ``Ask the other player to put something on top of something else'', or ``Question answering'', e.g. ``Ask the other player a question about the color of something''. 
 \textbf{C} Scripted probe tasks. The bottom right plot compares the success rate of humans and agents on six automated testing tasks. These supply a pre-scripted setter question or instruction (e.g. Position = ``Put the X near the Y''), where the response can be unambiguously scored as successful or unsuccessful by the environment engine. 
}
\label{fig:3}
\end{figure}

\agent{} achieves over $70\%$ success rate in human-rated online interactions, representing $75\%$ of that achieved by humans in similar positions. As indicated in figure \ref{fig:3}, according to our probe tasks \agent{} is most successful at ``going'' and ``lifting'', and less successful at more demanding motor tasks (such as positioning objects relative to each other) and language production tasks demanding complex cognition (such as counting). The results from our probe tasks also reveal weaknesses in scripting probe tasks for evaluation. For example, performance is low on tasks that ask the agent to mention the colors of items in the room, though this is not necessarily because the agent is poor at identifying colors. Rather, limitations in anticipating possible valid natural language responses (``blue'', ``light blue'', ``cyan'', ``sky blue'', etc.) prevent us from accurately assessing the model's abilities. Similar logic applies to motor tasks (e.g., how high should an agent lift an object to be ``successful''? What if the object is a table?). Indeed, post-hoc correction of our reward metric for ``counting'' can reveal a pre- and post- fluctuation of up to 30\% on the evaluated score (data not shown). It is precisely these researcher-centric deficiencies, if permeated into a reward model for reinforcement learning, that would result in less robust and less natural agent behaviours.

Figure \ref{fig:3} also indicates that we are in the over-fitted regime of training. The total loss shows mild overfitting---and the movement behavioural cloning loss perhaps none at all---but the language behavioural cloning and contrastive representation losses clearly indicate that agents could profit from more data. These results also highlight an aspect of integrated-agent training that is unique from more constrained domains like language modeling: the dynamics of each loss are different, making it difficult to apply techniques, such as model scaling, in a straightforward manner. While modeling motor actions could profit from larger models, this might be counteracted by more severe overfitting when modeling language. Nevertheless, a relatively simple way to improve performance (which we motivated further in section \ref{sec:figure_4}) is to collect and train on more data.

We performed a series of ablations and agent design modifications to better understand the role of various components in \agent{}. In particular, the ablations assessed the impact of removing visual inputs, language inputs, the auxiliary contrastive representation loss, and the hierarchical movement policy. We also determined the importance of visual resolution in the multi-modal transformer by modifying the structure of visual processing to downsample the image to $6\times5\times512$ (i.e., producing $30$-pixel feature maps as opposed to $432$-pixel feature maps in the baseline). 

As with many ablations, there is a confounding factor: removing agent components alters the agent's computational capacity (i.e., the number of total parameters), which can impact performance (see section \ref{sec:figure_4}). Unfortunately, there is no clear way to compensate for these parameter differences. However, as our data will show in section \ref{sec:figure_4}, since the differences in parameter counts for these ablations are minimal, performance loss due to decreased capacity should be minimal compared to the effects of the ablations themselves. The parameter counts for the various agents are: baseline $57$M, no language input $56$M, no vision $52$M, no auxiliary contrastive representation loss $56$M, no hierarchical movement policy $66$M, and modified visual processing to further downsample the input image $57$M. 

As the results show, there are clear performance degradations for each ablation and modification. Downsampling the input image less (i.e., using more pixels as input to the MMT) is best, though intriguingly, using a Vision Transformer-style technique~\citep{dosovitskiy2020image} that extracts image patches from the full resolution input image does not perform as well as using ResNet blocks (data not shown), which corroborates recent similar findings~\citep{wu2021cvt}.

While we provide evidence that the hierarchical movement policy significantly improves agent performance, the reason for its impact is unclear. We hypothesize that a few effects could be important: First, by forcing the agent to act every $8$ steps, we cause adjacent observations to be more visually distinct, which could aid learning (consider the converse scenario if the agent were forced to act with a higher frequency: adjacent observations would be nearly identical but have potentially different target policies, which creates complications for learning). Second, our implementation offers a form of data augmentation, as we can randomly vary the initial observation to be any of the first $8$ observations at the beginning of a given data trajectory. Third, the agent may genuinely, and implicitly learn useful hierarchical policies wherein its lower-level controller captures repeatable action sequences that merely need to be triggered by the higher-level controller.

Notably, when \agent{} is trained on the same Playroom data as in \cite{abramson2020imitating}, it outperforms the reported behavioural-cloning agent, which is augmented with privileged information, by $20\%$ on average across the single-room scripted probe tasks.

\section{Data and Parameter Scaling}
\label{sec:figure_4}
\begin{figure}[h]
    \centering
    \includegraphics[width=1\textwidth]{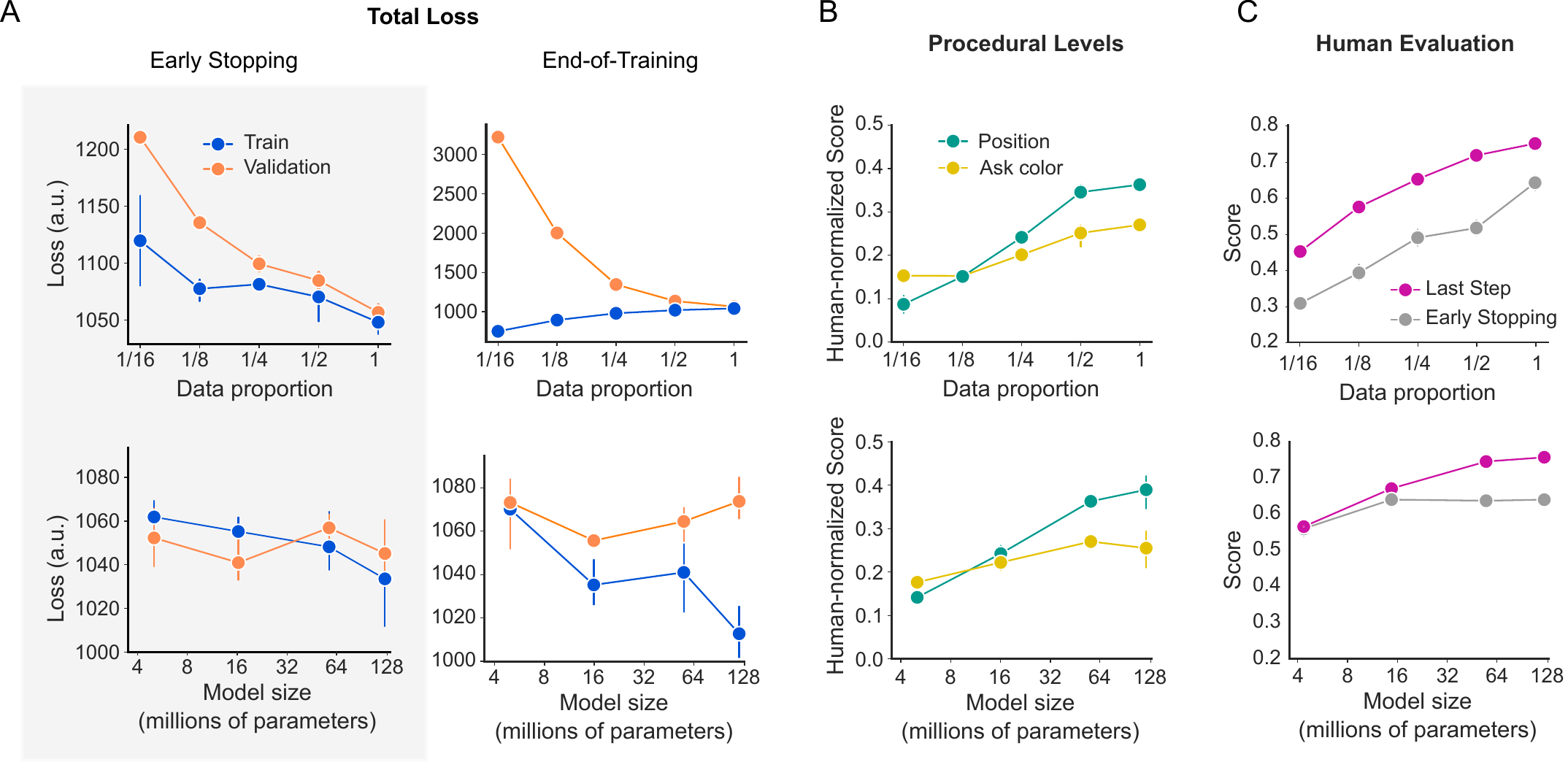}
    \caption{\textbf{Data and model scaling.} Total loss, scripted probe performance, and human evaluation for data and model scaling. All agents were trained to 5G steps, coinciding with convergence on scripted probe tasks, except for the early stopping condition which ends training at the point of lowest validation loss. Error bars represent standard error across 3 seeds. The baseline model with $56$M parameters was trained on differently-sized datasets (\textbf{A, B, C}. top row), or differently sized models were trained on the full dataset (\textbf{A, B, C}. bottom row). Validation losses decrease as dataset sizes increase, although this trend is not true for increasing model size (\textbf{A}). Nevertheless, for scripted probe tasks (\textbf{B}) and human evaluation (\textbf{C}) there are clear improvements when increasing both dataset size and model size. Performance on both scripted probe tasks (data not shown) and human evaluation (\textbf{C}) is worse at the early stopping point, without exception.
}
\label{fig:4}
\end{figure}

Contemporary machine learning research has uncovered remarkable empirical effects regarding scale~\citep[e.g.]{kaplan2020scaling}; in particular, model performance scales using a power-law trend with dataset size, model size (in terms of number of parameters), and compute. These effects have traditionally been observed in the language domain, which is characterized by massive dataset sizes, standardized architectures, and refined training protocols. In this work, however, we are in a decidedly different regime. We assume we have collected an abundance of data, but the results in figure \ref{fig:4}, particularly the losses, indicate that we could still be in a ``low data'' regime where overfitting is more pronounced and scaling effects are more difficult to see~\citep{kaplan2020scaling}. Moreover, the cognitively rich tasks force us to iterate our model architecture as we explore the effects of our choices on, for example, memory formation and retrieval, language production, motor control, and visual object identification, making model inductive biases a perpetually confounding factor in our research. This architectural flux, combined with a complicated training protocol involving a handful of losses to optimize, lead to natural questions of whether scaling effects can also be observed in the Playhouse.

In figure \ref{fig:4} we present some mixed evidence after training four differently-sized models ($5$M, $16$M, $56$M (baseline), and $121$M) on five differently-sized datasets ($\frac{1}{16}$, $\frac{1}{8}$, $\frac{1}{4}$, $\frac{1}{2}$, and the full dataset size). In terms of the total training loss, larger models achieve lower values for each dataset size, and predictably, training losses are lower in the smaller data regimes (implying a easier ability to memorize particular examples as dataset sizes decrease). The effects of scaling on the validation loss are more difficult to interpret: the total loss shows little difference between model sizes, though it displays a clear trend with dataset size when keeping model size constant. When analyzing individual losses we see a clear effect for some, such as the hierarchical movement BC loss. The language and contrastive representation losses, on the other hand, show clear and quick overfitting for each model (as in figure \ref{fig:2}). 

To get a clearer picture we also measured the performance metric that we ultimately care about: human-evaluated behavioural competency during online interactions. The results here are consistent (low standard error across seeds) and clear: increasing model size and dataset size clearly benefit performance. 

Altogether, these results lead us to believe that we may be in the low-data regime for the Playhouse, suggesting that performance gains are to be had by collecting more data (and then subsequently scaling model size). The results also highlight the problems with analyzing the effects scaling in these more complicated settings where a handful of losses, jointly optimized, each exhibit different learning dynamics and interact with each other in unknown ways: the values we observe for the losses may obfuscate the effects of scale on ultimately more important \textit{behavioural} metrics.
\section{Behavioural Transfer}
\begin{figure}[h]
    \centering
    \includegraphics[width=\textwidth]{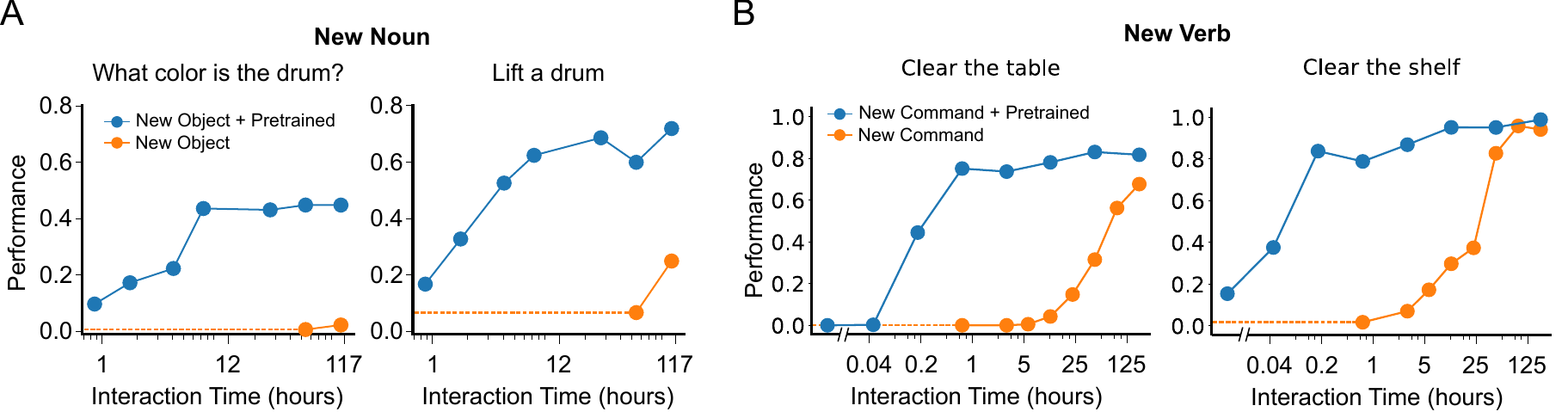}
    \caption{\textbf{Behavioural Transfer.} We quantified the agent performance when learning a new noun (a drum) or verb (to clear) as a function of quantity of human demonstrations, measured in hours. With both a new noun and new verb agent performance quickly improves with mere hours of demonstration experience.}
    \label{fig:5}
\end{figure}

We investigated how much data is needed to learn how to interact with a new, previously unseen object (new noun). Starting with a pre-trained \agent{}, we began new training on varying amounts of data from humans interacting with a novel object (a drum), in addition to the data from the original dataset. The amount of data is quantified in hours of new data measured as real time experience (note: this measures the time required to collect this data in human hours, not the amount of time the agent experiences it during the course of supervised training). After training, we evaluated the ability of the agents to perform simple tasks with the new object (answering about its color and lifting it). We observe that using less than $12$ hours of human interaction is enough to reach the final performance on subsequent language and motor tasks involving drums. (figure \ref{fig:5} B). 

To study how much data is needed to learn a new command (the verb ``to clear''), we again start with a pre-trained \agent{} and co-train it on novel episodes where humans are instructed to remove all objects from a surface (e.g., table or shelf). The original pre-trained agent is incapable of performing this behavior, which is to be expected as this command is not present in the original data. However, after training on novel episodes containing examples of this behavior, we observe that with around $1$ hour of human demonstrations the agent reaches near optimal performance at this task.

Altogether these results suggest that while agents do not generalize zero-shot to new nouns or verbs, they have a sufficiently rich behavioural prior such that subsequent training on small amounts of experience allow rapid adaptation to new objects and actions.
\section{Discussion \& Conclusion}
In this work we've built upon the paradigm introduced in \cite{abramson2020imitating} to construct an agent, \agent, that emphasises a holistic integration of embodied control, perception, and language understanding and production~\citep{mcclelland2019extending,lake2021word}. The goal was to elicit behaviours that are naturalistic and appropriately exploratory~\citep{turing,winograd1972understanding}, and hence, are capable of subsequent efficient tuning using reinforcement. 

Imitation of human behaviour was central to our training approach. Unlike other domains, like language~\citep{devlin2018bert,radford2019language,brown2020language}, human behavioural data in 3D simulated worlds often needs to be collected and curated from scratch. Some previous work has collected human (child) multimodal behavioural data~\citep{roy2006human,yoshida2008s,sullivan2020saycam}, but has not used it to train artificial agents. 
Social learning, imitation, and mimicry is prevalent in the the animal kingdom~\citep{laland2004social,byrne2009animal}. In humans, infants naturally imitate both the language and actions they encounter~\citep{chomsky1959chomsky,heyes1996social}. However, in AI, imitating multi-modal, embodied behavioural data is especially challenging because latent in all trajectories of human experience are intentions and goals. It is not clear whether we currently possess the algorithms and models that can interpret implicit causal factors of human behaviour~\citep{ortega2021shaking}, and hence, can engage in consistent and coherent naturalistic interactions. Nevertheless, as in previous work we observe that some techniques, such as dataset and model scaling, reliably improve performance as measured by human evaluators. In addition, a number of architectural improvements proved crucial: self-supervised learning that gauged whether cross-modality representations ``matched'' or not, and hierarchical control of motor actions both substantially improved agent performance. 

Our setting is a natural one in which to study ``language grounding''~\citep{harnad1990symbol}, wherein sensorimotor embodiment combines with language production and understanding to produce models that more closely approach human-like symbolic behaviour~\citep{santoro2021symbolic}. Work in robotics and in 3D simulated environments has pursued similar ideas, such as  natural language conditioning of tasks~\citep{lynch2020grounding,hill2019robust,DBLP:journals/corr/abs-1711-07280,das2018embodied}. However, to our knowledge our work is unique in combining language conditioning with unconstrained language production, in an embodied setting demanding complex navigation and manipulation for upwards of $5$ minutes, with a human in-the-loop.

An exciting direction that remains unexplored is to also train setter agents using the data produced by humans whose role it is to set tasks and questions. ``Self-play''-like settings have proved powerful in developing agents in games, where agent roles are symmetric, and reward signals provide positive and negative feedback signals per experience~\citep{silver2016mastering,vinyals2019grandmaster}. It is less clear, however, how one could leverage agent setter-agent solver dynamics in the Playhouse environment where roles are asymmetric and rewards do not exist, beyond using the extra data for generic representation learning.

One challenge for future work pertains to the diversity and flexibility of language comprehended or produced by our agents, as it is far more constrained (both semantically and syntactically) than in the case of non-embodied systems trained on web-scale text data~\citep{brown2020language}. Nevertheless, a useful intelligent robot would not need to be an unconstrained source of knowledge, provided that it can communicate efficiently and effectively about its physical environment.

In addition, the control and perception challenges faced by our agents are substantially simpler than they would be in a physical robot. While our data is generated by participants sitting at computers, making research comparatively rapid, analogous data for a physically-embodied interactive agent might require teleoperation of specialized and expensive hardware~\citep{jang2021bc}. On the other hand, unlike many current robotic systems, our agent must integrate learning perception and control with memory and language. Generic models for perception and control skill learning are rapidly improving, and we hope that such techniques might soon combine with the approach described here to yield a system capable of all of these things. Another key challenge when transitioning from simulation to reality is the collection of adequate data. 

A final pressing challenge for future work is how to appropriately evaluate interactive agents like \agent. None of our current mechanisms--- training loss, scripted probe tasks, and online human-agent evaluation---are ideal for research. Training losses and scripted probe tasks are only heuristic measures of performance; they generally correlate with better agents, but these metrics do not necessarily move monotonically with human judgement. This is particularly troublesome since we explicitly optimize the training loss, and not human judgement, and ideally the measure that we optimize should be precisely the measure that we ultimately care about. On the other hand, real-time human-agent evaluation is expensive, requiring many human hours. It is also high-variance and inconsistent, as live human interactions will necessarily be different every time, and humans may shift their content over time. Nevertheless, models trained to mimic human judgements may ultimately help us tackle open-ended evaluation for multi-modal and real-time environments.

\newpage
\section{Authors \& Contributions}

\textbf{Josh Abramson} contributed to agent development, imitation learning, data and tasks, running and analysis of experiments, engineering infrastructure, writing, and as a technical lead. \\
\textbf{Arun Ahuja} contributed to agent development, imitation learning, data and tasks, running and analysis of experiments, engineering infrastructure, writing, and as a technical lead. \\
\textbf{Arthur Brussee} contributed to environment development. \\
\textbf{Federico Carnevale} contributed to agent development, imitation learning, running and analysis of experiments, writing, and as a sub-effort lead for agent development. \\
\textbf{Mary Cassin} contributed to environment development. \\
\textbf{Felix Fischer} contributed to engineering infrastructure. \\
\textbf{Petko Georgiev} contributed to agent development, engineering infrastructure, evaluation development, environment development, running and analysis of experiments and as a technical lead. \\
\textbf{Alex Goldin} contributed to project management. \\
\textbf{Mansi Gupta} contributed to engineering infrastructure. \\
\textbf{Tim Harley} contributed to engineering infrastructure. \\
\textbf{Felix Hill} contributed to environment development. \\
\textbf{Peter C Humphreys} contributed to agent development and writing. \\
\textbf{Alden Hung} contributed to agent development, imitation learning and as a sub-effort lead for agent development. \\
\textbf{Jessica Landon} contributed to data and tasks, engineering infrastructure, evaluation development and as a sub-effort lead for data and evaluation. \\
\textbf{Timothy Lillicrap} contributed to agent development, imitation learning, data and tasks, environment development, evaluation development, writing, and as an effort lead. \\
\textbf{Hamza Merzic} contributed to technical infrastructure. \\
\textbf{Alistair Muldal} contributed to data and tasks, evaluation development and as a sub-effort lead for data and evaluation. \\
\textbf{Adam Santoro} contributed to agent development, imitation learning, running and analysis of experiments and writing.\\
\textbf{Guy Scully} contributed to project management. \\
\textbf{Tamara von Glehn} contributed to agent development, engineering infrastructure, imitation learning and running and analysis of experiments. \\
\textbf{Greg Wayne} contributed to agent development, imitation learning, data and tasks, environment development, evaluation development, writing, and as an effort lead. \\
\textbf{Nathaniel Wong} contributed to environment development. \\
\textbf{Chen Yan} contributed to agent development, data and tasks, engineering infrastructure, imitation learning, running and analysis of experiments and writing. \\
\textbf{Rui Zhu} contributed to agent development, engineering infrastructure, environment development and writing. \\

\vspace{1mm}
\noindent
{\bf Corresponding Authors:} \\
Greg Wayne (gregwayne@deepmind.com) \& Timothy Lillicrap (countzero@deepmind.com)

\section{Acknowledgments}

The authors would like to thank Duncan Williams, Daan Wierstra, Dario de Cesare, Koray Kavukcuoglu, Matt Botvinick, Lorrayne Bennett, the Worlds Team, and Crowd Compute.
\bibliographystyle{unsrtnat}
\bibliography{references}

\begin{thebibliography}{40}
\providecommand{\natexlab}[1]{#1}
\providecommand{\url}[1]{\texttt{#1}}
\expandafter\ifx\csname urlstyle\endcsname\relax
  \providecommand{\doi}[1]{doi: #1}\else
  \providecommand{\doi}{doi: \begingroup \urlstyle{rm}\Url}\fi

\bibitem[Dunbar(1993)]{dunbar1993coevolution}
Robin~IM Dunbar.
\newblock Coevolution of neocortical size, group size and language in humans.
\newblock \emph{Behavioral and brain sciences}, 16\penalty0 (4):\penalty0
  681--694, 1993.

\bibitem[McClelland et~al.(2019)McClelland, Hill, Rudolph, Baldridge, and
  Sch{\"u}tze]{mcclelland2019extending}
James~L McClelland, Felix Hill, Maja Rudolph, Jason Baldridge, and Hinrich
  Sch{\"u}tze.
\newblock Extending machine language models toward human-level language
  understanding.
\newblock \emph{arXiv preprint arXiv:1912.05877}, 2019.

\bibitem[Lake and Murphy(2021)]{lake2021word}
Brenden~M Lake and Gregory~L Murphy.
\newblock Word meaning in minds and machines.
\newblock \emph{Psychological Review}, 2021.

\bibitem[Winograd(1972)]{winograd1972understanding}
Terry Winograd.
\newblock Understanding natural language.
\newblock \emph{Cognitive psychology}, 3\penalty0 (1):\penalty0 1--191, 1972.

\bibitem[Pomerleau(1989)]{pomerleau1989alvinn}
Dean~A Pomerleau.
\newblock Alvinn: An autonomous land vehicle in a neural network.
\newblock In \emph{Advances in neural information processing systems}, pages
  305--313, 1989.

\bibitem[Schaal(1999)]{schaal1999imitation}
Stefan Schaal.
\newblock Is imitation learning the route to humanoid robots?
\newblock \emph{Trends in cognitive sciences}, 3\penalty0 (6):\penalty0
  233--242, 1999.

\bibitem[Silver et~al.(2016)Silver, Huang, Maddison, Guez, Sifre, Van
  Den~Driessche, Schrittwieser, Antonoglou, Panneershelvam, Lanctot,
  et~al.]{silver2016mastering}
David Silver, Aja Huang, Chris~J Maddison, Arthur Guez, Laurent Sifre, George
  Van Den~Driessche, Julian Schrittwieser, Ioannis Antonoglou, Veda
  Panneershelvam, Marc Lanctot, et~al.
\newblock Mastering the game of go with deep neural networks and tree search.
\newblock \emph{nature}, 529\penalty0 (7587):\penalty0 484--489, 2016.

\bibitem[Vinyals et~al.(2019)Vinyals, Babuschkin, Czarnecki, Mathieu, Dudzik,
  Chung, Choi, Powell, Ewalds, Georgiev, et~al.]{vinyals2019grandmaster}
Oriol Vinyals, Igor Babuschkin, Wojciech~M Czarnecki, Micha{\"e}l Mathieu,
  Andrew Dudzik, Junyoung Chung, David~H Choi, Richard Powell, Timo Ewalds,
  Petko Georgiev, et~al.
\newblock Grandmaster level in starcraft ii using multi-agent reinforcement
  learning.
\newblock \emph{Nature}, 575\penalty0 (7782):\penalty0 350--354, 2019.

\bibitem[Brown et~al.(2020)Brown, Mann, Ryder, Subbiah, Kaplan, Dhariwal,
  Neelakantan, Shyam, Sastry, Askell, et~al.]{brown2020language}
Tom~B Brown, Benjamin Mann, Nick Ryder, Melanie Subbiah, Jared Kaplan, Prafulla
  Dhariwal, Arvind Neelakantan, Pranav Shyam, Girish Sastry, Amanda Askell,
  et~al.
\newblock Language models are few-shot learners.
\newblock \emph{arXiv preprint arXiv:2005.14165}, 2020.

\bibitem[Adiwardana et~al.(2020)Adiwardana, Luong, So, Hall, Fiedel, Thoppilan,
  Yang, Kulshreshtha, Nemade, Lu, et~al.]{adiwardana2020towards}
Daniel Adiwardana, Minh-Thang Luong, David~R So, Jamie Hall, Noah Fiedel, Romal
  Thoppilan, Zi~Yang, Apoorv Kulshreshtha, Gaurav Nemade, Yifeng Lu, et~al.
\newblock Towards a human-like open-domain chatbot.
\newblock \emph{arXiv preprint arXiv:2001.09977}, 2020.

\bibitem[Interactive Agents~Team(2020)]{abramson2020imitating}
DeepMind Interactive Agents~Team.
\newblock Imitating interactive intelligence.
\newblock \emph{arXiv preprint arXiv:2012.05672}, 2020.

\bibitem[Ward et~al.(2020)Ward, Bolt, Hemmings, Carter, Sanchez, Barreira,
  Noury, Anderson, Lemmon, Coe, et~al.]{ward2020using}
Tom Ward, Andrew Bolt, Nik Hemmings, Simon Carter, Manuel Sanchez, Ricardo
  Barreira, Seb Noury, Keith Anderson, Jay Lemmon, Jonathan Coe, et~al.
\newblock Using unity to help solve intelligence.
\newblock \emph{arXiv preprint arXiv:2011.09294}, 2020.

\bibitem[Christiano et~al.(2017)Christiano, Leike, Brown, Martic, Legg, and
  Amodei]{christiano2017deep}
Paul Christiano, Jan Leike, Tom~B Brown, Miljan Martic, Shane Legg, and Dario
  Amodei.
\newblock Deep reinforcement learning from human preferences.
\newblock \emph{arXiv preprint arXiv:1706.03741}, 2017.

\bibitem[Galashov et~al.(2019)Galashov, Jayakumar, Hasenclever, Tirumala,
  Schwarz, Desjardins, Czarnecki, Teh, Pascanu, and
  Heess]{galashov2019information}
Alexandre Galashov, Siddhant~M Jayakumar, Leonard Hasenclever, Dhruva Tirumala,
  Jonathan Schwarz, Guillaume Desjardins, Wojciech~M Czarnecki, Yee~Whye Teh,
  Razvan Pascanu, and Nicolas Heess.
\newblock Information asymmetry in kl-regularized rl.
\newblock \emph{arXiv preprint arXiv:1905.01240}, 2019.

\bibitem[Ross et~al.(2011)Ross, Gordon, and Bagnell]{ross2011reduction}
St{\'e}phane Ross, Geoffrey Gordon, and Drew Bagnell.
\newblock A reduction of imitation learning and structured prediction to
  no-regret online learning.
\newblock In \emph{Proceedings of the fourteenth international conference on
  artificial intelligence and statistics}, pages 627--635. JMLR Workshop and
  Conference Proceedings, 2011.

\bibitem[Lynch and Sermanet(2020)]{lynch2020grounding}
Corey Lynch and Pierre Sermanet.
\newblock Grounding language in play.
\newblock \emph{arXiv preprint arXiv:2005.07648}, 2020.

\bibitem[Osa et~al.(2018)Osa, Pajarinen, Neumann, Bagnell, Abbeel, and
  Peters]{osa2018algorithmic}
Takayuki Osa, Joni Pajarinen, Gerhard Neumann, J~Andrew Bagnell, Pieter Abbeel,
  and Jan Peters.
\newblock An algorithmic perspective on imitation learning.
\newblock \emph{arXiv preprint arXiv:1811.06711}, 2018.

\bibitem[Alayrac et~al.(2020)Alayrac, Recasens, Schneider, Arandjelovic,
  Ramapuram, De~Fauw, Smaira, Dieleman, and Zisserman]{alayrac2020self}
Jean-Baptiste Alayrac, Adria Recasens, Rosalia Schneider, Relja Arandjelovic,
  Jason Ramapuram, Jeffrey De~Fauw, Lucas Smaira, Sander Dieleman, and Andrew
  Zisserman.
\newblock Self-supervised multimodal versatile networks.
\newblock \emph{NeurIPS}, 2\penalty0 (6):\penalty0 7, 2020.

\bibitem[He et~al.(2016)He, Zhang, Ren, and Sun]{he2016deep}
Kaiming He, Xiangyu Zhang, Shaoqing Ren, and Jian Sun.
\newblock Deep residual learning for image recognition.
\newblock In \emph{Proceedings of the IEEE conference on computer vision and
  pattern recognition}, pages 770--778, 2016.

\bibitem[Devlin et~al.(2018)Devlin, Chang, Lee, and Toutanova]{devlin2018bert}
Jacob Devlin, Ming-Wei Chang, Kenton Lee, and Kristina Toutanova.
\newblock Bert: Pre-training of deep bidirectional transformers for language
  understanding.
\newblock \emph{arXiv preprint arXiv:1810.04805}, 2018.

\bibitem[Vaswani et~al.(2017)Vaswani, Shazeer, Parmar, Uszkoreit, Jones, Gomez,
  Kaiser, and Polosukhin]{vaswani2017attention}
Ashish Vaswani, Noam Shazeer, Niki Parmar, Jakob Uszkoreit, Llion Jones,
  Aidan~N Gomez, {\L}ukasz Kaiser, and Illia Polosukhin.
\newblock Attention is all you need.
\newblock In \emph{Advances in neural information processing systems}, pages
  5998--6008, 2017.

\bibitem[Dosovitskiy et~al.(2020)Dosovitskiy, Beyer, Kolesnikov, Weissenborn,
  Zhai, Unterthiner, Dehghani, Minderer, Heigold, Gelly,
  et~al.]{dosovitskiy2020image}
Alexey Dosovitskiy, Lucas Beyer, Alexander Kolesnikov, Dirk Weissenborn,
  Xiaohua Zhai, Thomas Unterthiner, Mostafa Dehghani, Matthias Minderer, Georg
  Heigold, Sylvain Gelly, et~al.
\newblock An image is worth 16x16 words: Transformers for image recognition at
  scale.
\newblock \emph{arXiv preprint arXiv:2010.11929}, 2020.

\bibitem[Wu et~al.(2021)Wu, Xiao, Codella, Liu, Dai, Yuan, and
  Zhang]{wu2021cvt}
Haiping Wu, Bin Xiao, Noel Codella, Mengchen Liu, Xiyang Dai, Lu~Yuan, and Lei
  Zhang.
\newblock Cvt: Introducing convolutions to vision transformers.
\newblock \emph{arXiv preprint arXiv:2103.15808}, 2021.

\bibitem[Kaplan et~al.(2020)Kaplan, McCandlish, Henighan, Brown, Chess, Child,
  Gray, Radford, Wu, and Amodei]{kaplan2020scaling}
Jared Kaplan, Sam McCandlish, Tom Henighan, Tom~B Brown, Benjamin Chess, Rewon
  Child, Scott Gray, Alec Radford, Jeffrey Wu, and Dario Amodei.
\newblock Scaling laws for neural language models.
\newblock \emph{arXiv preprint arXiv:2001.08361}, 2020.

\bibitem[Turing(1950)]{turing}
Alan~M Turing.
\newblock Computing machinery and intelligence.
\newblock \emph{Mind}, LIX\penalty0 (236):\penalty0 433–460, 1950.

\bibitem[Radford et~al.(2019)Radford, Wu, Child, Luan, Amodei, and
  Sutskever]{radford2019language}
Alec Radford, Jeffrey Wu, Rewon Child, David Luan, Dario Amodei, and Ilya
  Sutskever.
\newblock Language models are unsupervised multitask learners.
\newblock \emph{OpenAI Blog}, 1\penalty0 (8):\penalty0 9, 2019.

\bibitem[Roy et~al.(2006)Roy, Patel, DeCamp, Kubat, Fleischman, Roy, Mavridis,
  Tellex, Salata, Guinness, et~al.]{roy2006human}
Deb Roy, Rupal Patel, Philip DeCamp, Rony Kubat, Michael Fleischman, Brandon
  Roy, Nikolaos Mavridis, Stefanie Tellex, Alexia Salata, Jethran Guinness,
  et~al.
\newblock The human speechome project.
\newblock In \emph{International Workshop on Emergence and Evolution of
  Linguistic Communication}, pages 192--196. Springer, 2006.

\bibitem[Yoshida and Smith(2008)]{yoshida2008s}
Hanako Yoshida and Linda~B Smith.
\newblock What's in view for toddlers? using a head camera to study visual
  experience.
\newblock \emph{Infancy}, 13\penalty0 (3):\penalty0 229--248, 2008.

\bibitem[Sullivan et~al.(2020)Sullivan, Mei, Perfors, Wojcik, and
  Frank]{sullivan2020saycam}
Jess Sullivan, Michelle Mei, Amy Perfors, Erica~H Wojcik, and Michael~C Frank.
\newblock Saycam: A large, longitudinal audiovisual dataset recorded from the
  infant’s perspective.
\newblock \emph{PsyArXiv}, 2020.

\bibitem[Laland(2004)]{laland2004social}
Kevin~N Laland.
\newblock Social learning strategies.
\newblock \emph{Animal Learning \& Behavior}, 32\penalty0 (1):\penalty0 4--14,
  2004.

\bibitem[Byrne(2009)]{byrne2009animal}
Richard~W Byrne.
\newblock Animal imitation.
\newblock \emph{Current Biology}, 19\penalty0 (3):\penalty0 R111--R114, 2009.

\bibitem[Chomsky(1959)]{chomsky1959chomsky}
Noam Chomsky.
\newblock Chomsky, n. 1959. a review of bf skinner’s verbal behavior.
  language, 35 (1), 26--58., 1959.

\bibitem[Heyes and Galef~Jr(1996)]{heyes1996social}
Cecilia~M Heyes and Bennett~G Galef~Jr.
\newblock \emph{Social learning in animals: the roots of culture}.
\newblock Elsevier, 1996.

\bibitem[Ortega et~al.(2021)Ortega, Kunesch, Del{\'e}tang, Genewein, Grau-Moya,
  Veness, Buchli, Degrave, Piot, Perolat, et~al.]{ortega2021shaking}
Pedro~A Ortega, Markus Kunesch, Gr{\'e}goire Del{\'e}tang, Tim Genewein, Jordi
  Grau-Moya, Joel Veness, Jonas Buchli, Jonas Degrave, Bilal Piot, Julien
  Perolat, et~al.
\newblock Shaking the foundations: delusions in sequence models for interaction
  and control.
\newblock \emph{arXiv preprint arXiv:2110.10819}, 2021.

\bibitem[Harnad(1990)]{harnad1990symbol}
Stevan Harnad.
\newblock The symbol grounding problem.
\newblock \emph{Physica D: Nonlinear Phenomena}, 42\penalty0 (1-3):\penalty0
  335--346, 1990.

\bibitem[Santoro et~al.(2021)Santoro, Lampinen, Mathewson, Lillicrap, and
  Raposo]{santoro2021symbolic}
Adam Santoro, Andrew Lampinen, Kory Mathewson, Timothy Lillicrap, and David
  Raposo.
\newblock Symbolic behaviour in artificial intelligence.
\newblock \emph{arXiv preprint arXiv:2102.03406}, 2021.

\bibitem[Hill et~al.(2019)Hill, Mokra, Wong, and Harley]{hill2019robust}
Felix Hill, Sona Mokra, Nathaniel Wong, and Tim Harley.
\newblock Robust instruction-following in a situated agent via
  transfer-learning from text.
\newblock \emph{OpenReview}, 2019.

\bibitem[Anderson et~al.(2017)Anderson, Wu, Teney, Bruce, Johnson,
  S{\"{u}}nderhauf, Reid, Gould, and van~den
  Hengel]{DBLP:journals/corr/abs-1711-07280}
Peter Anderson, Qi~Wu, Damien Teney, Jake Bruce, Mark Johnson, Niko
  S{\"{u}}nderhauf, Ian~D. Reid, Stephen Gould, and Anton van~den Hengel.
\newblock Vision-and-language navigation: Interpreting visually-grounded
  navigation instructions in real environments.
\newblock \emph{CoRR}, abs/1711.07280, 2017.
\newblock URL \url{http://arxiv.org/abs/1711.07280}.

\bibitem[Das et~al.(2018)Das, Datta, Gkioxari, Lee, Parikh, and
  Batra]{das2018embodied}
Abhishek Das, Samyak Datta, Georgia Gkioxari, Stefan Lee, Devi Parikh, and
  Dhruv Batra.
\newblock Embodied question answering.
\newblock In \emph{Proceedings of the IEEE Conference on Computer Vision and
  Pattern Recognition Workshops}, pages 2054--2063, 2018.

\bibitem[Jang et~al.(2021)Jang, Irpan, Khansari, Kappler, Ebert, Lynch, Levine,
  and Finn]{jang2021bc}
Eric Jang, Alex Irpan, Mohi Khansari, Daniel Kappler, Frederik Ebert, Corey
  Lynch, Sergey Levine, and Chelsea Finn.
\newblock Bc-0: Zero-shot task generalization with robotic imitation learning.
\newblock In \emph{5th Annual Conference on Robot Learning}, 2021.

\end{thebibliography}
\end{document}